\documentclass[10pt,twocolumn,letterpaper]{article}

\usepackage{cvpr}              
\usepackage{multirow}
\usepackage{makecell}
\usepackage{float}
\usepackage{comment}
\usepackage[dvipsnames]{xcolor}

\definecolor{cvprblue}{rgb}{0.21,0.49,0.74}
\usepackage[pagebackref,breaklinks,colorlinks,citecolor=cvprblue]{hyperref}

\def\paperID{*****} 
\def\paperID{840}
\def\confName{CVPR}
\def\confYear{2024}

\newcommand{\name}{\textsc{AV-Conv}\xspace}

\title{The Audio-Visual Conversational Graph: \\ From an Egocentric-Exocentric Perspective}

\author{Wenqi Jia$^{1,2,\ast}$, Miao Liu$^4$, Hao Jiang$^2$, Ishwarya Ananthabhotla$^2$, \\James M. Rehg$^3$, Vamsi Krishna Ithapu$^2$, Ruohan Gao$^2$\\
$^1$Georgia Tech, $^2$Meta Reality Labs, $^3$UIUC, $^4$GenAI, Meta\\
{\tt\small wenqi.jia@gatech.edu, \{miaoliu,haojiang,ishwarya,ithapu,rhgao\}@meta.com, jrehg@illinois.edu}\\
}\date{}

\begin{document}
\maketitle
\begin{abstract}

In recent years, the thriving development of research related to egocentric videos has provided a unique perspective for the study of conversational interactions, where both visual and audio signals play a crucial role. While most prior work focus on learning about behaviors that directly involve the camera wearer, we introduce the Ego-Exocentric Conversational Graph Prediction problem, marking the first attempt to infer exocentric conversational interactions from egocentric videos. We propose a unified multi-modal framework---Audio-Visual Conversational Attention (\name), for the joint prediction of conversation behaviors---speaking and listening---for both the camera wearer as well as all other social partners present in the egocentric video. Specifically, we adopt the self-attention mechanism to model the representations across-time, across-subjects, and across-modalities. To validate our method, we conduct experiments on a challenging egocentric video dataset that includes multi-speaker and multi-conversation scenarios. Our results demonstrate the superior performance of our method compared to a series of baselines. We also present detailed ablation studies to assess the contribution of each component in our model. Check our \href{https://vjwq.github.io/AV-CONV/}{Project Page}.
\let\thefootnote\relax\footnotetext{* This work was done during an internship at Meta Reality Labs.}

\end{abstract}    
\section{Introduction}
\label{sec:intro}




\begin{figure}[t]
\centering
\includegraphics[width=1.0\linewidth]{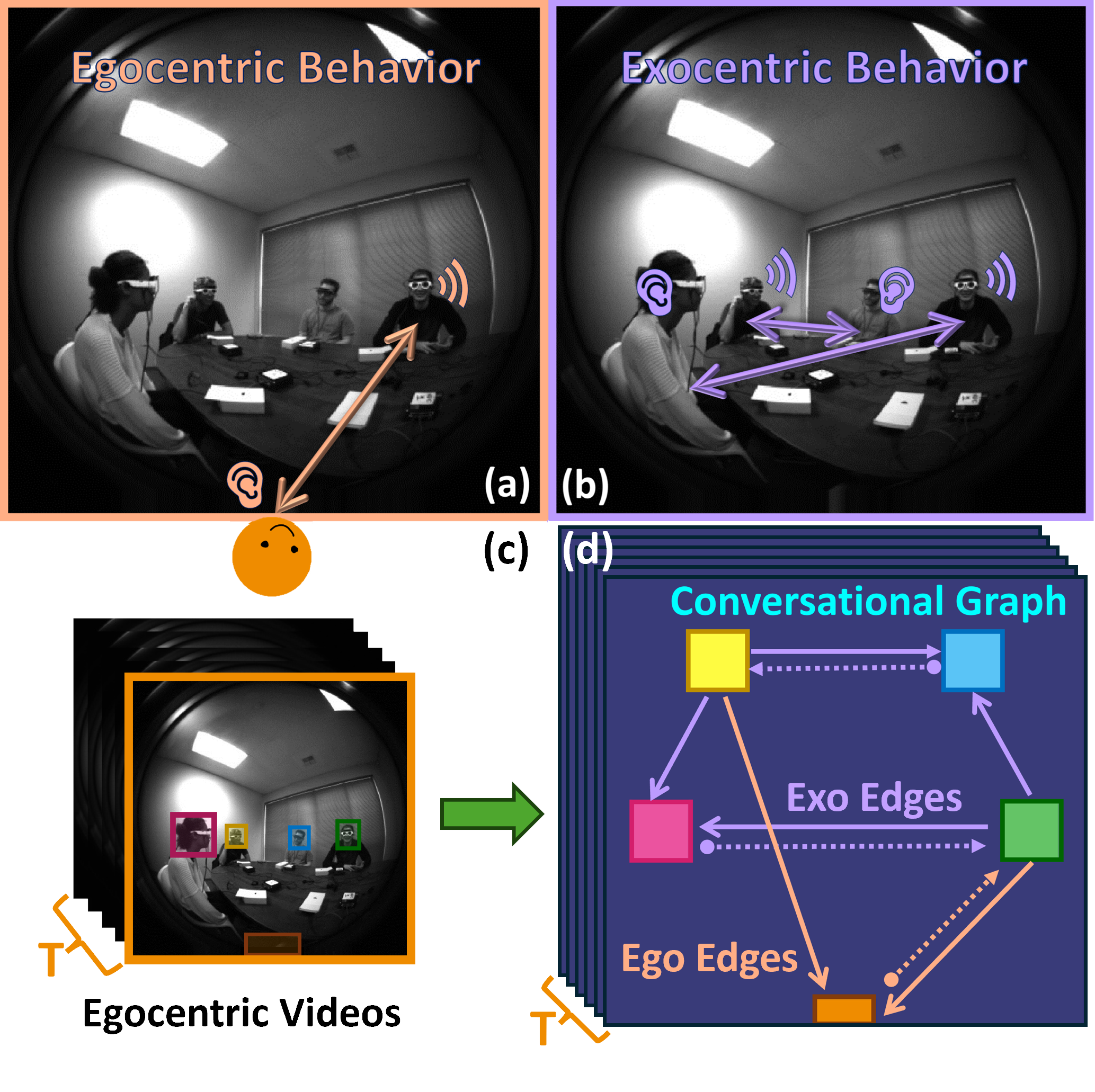}
\caption{We propose (d) the Ego-Exocentric Conversational Graph Prediction problem that jointly learns (a) the egocentric behaviors---whether the camera wearer is speaking or listening to others, and (b) the exocentric behaviors---whether the other social partners in the scene are speaking or listening to one another, given only the egocentric video input (c).}
\label{fig:teaser}
\end{figure}

With the thriving growth of social media and message-based communication in recent decades, the concept of \textit{Social Graph}~\cite{ugander2011anatomy} has found wide applications, such as on human intention understanding, personalized recommendations~\cite{granovetter1983strength, leskovec2012learning, perry2018egocentric}, etc. Going beyond texts and images, the latest advancement of VR/AR technology and egocentric perception presents a new multimodal data format and also the associated new challenges on constructing graph representation for the multi-participant conversational behaviors in egocentric video streams.

While there has been exciting progress on building large-scale egocentric datasets~\cite{damen2018scaling,grauman2022ego4d} and models to detect the talking and listening behaviors \cite{northcutt2020egocom, jiang2022egocentric, ryan2023egocentric} of the subjects in egocentric videos, existing work mostly focus on analyzing behaviors or actions that directly involve the camera wearer (ego). However, cognitive science studies tell us that we humans also have remarkable ability of understanding other people' belief state during social interactions, often referred as Theory of Mind (ToM)~\cite{wellman1992child}. Subjects involved in a conversation often enact orienting behaviors, such as shifting the position of the head or the focus of the eyes and other overt behaviors, such as  hand gestures, synchrony in movement, all of which can be indicators of their explicit (i.e., actively engaged in a conversation) or implicit (i.e., listening discreetly) visual and auditory attention \cite{frischen2007gaze, brimijoin2010auditory, lu2022sound}.

Motivated by the above, we introduce the concept of \emph{Audio-Visual Conversational Graph}, which describes the conversational behaviors---speaking and listening---for both the camera wearer and all social partners involved in the conversation. As shown in Fig.~\ref{fig:teaser}, we present a challenging Ego-Exocentric Conversational Graph Prediction problem.
When provided with an egocentric video containing multiple people actively engaged in a conversation, our goal is to create a complete dynamic directed graph that can instantly reflect the conversational behaviors and relationships among all participants. Unlike prior work that focuses on only a single task~\cite{jiang2022egocentric,ryan2023egocentric, jia2022generative}, we simultaneously address multiple closely related tasks in a unified framework, and we are the first to explicitly predict the dense conversational behaviors from an exocentric point of view. 

We propose the Audio-Visual Conversational Attention (\name) model that leverages both the multi-channel audio and visual information for analyzing the behaviors and relationships between different social partners. We use a self-attention mechanism tailored to the egocentric conversation setting, and fuse information across-time, across-subjects, and across-modalities. We evaluate our model on a complex multi-speaker, multi-conversation dataset, and obtain an average accuracy of 86.15\% on egocentric-related predictions, and an average accuracy of 81.04\% on exocentric-related predictions, significantly outperforming the baseline methods.
In summary, we make the following main contributions:
\begin{itemize}
    \vspace{0.05in}
    \item We introduce the Ego-Exocentric Conversational Graph Prediction problem, the first attempt to explore exocentric conversational interactions from egocentric videos.
    \vspace{0.05in}
    \item We propose a unified multi-modal framework that jointly predicts the interaction states of all social entities captured in the egocentric videos. 
    \vspace{0.05in}
    \item Evaluating our \name model on a challenging first-person perspective multi-speaker, multi-conversation dataset, we demonstrate the effectiness of our model design compared to baseline methods. 
\end{itemize}

\vspace{0.2in}
\section{Related Work}
\label{sec:related}
\noindent\textbf{Exocentric Conversation Interaction}.\ 
Modeling the interactive behavior between people in a conversation has been a challenging problem. The term \textit{F-formations}, first introduced by anthropologist Edward Hall ~\cite{hall1966hidden} and further defined by Ciolek and Kendon~\cite{ciolek1980environment}, represents the spatial arrangement of individuals in a group during face-to-face communication. In recent years, a series of work from the computer vision community have contributed to this topic~\cite{hung2011detecting, cristani2011social, ricci2015uncovering, vascon2016detecting, hedayati2019recognizing, tan2022conversation}. However, these studies focus on monitoring crowd activities from a surveillance camera or overhead camera perspective, and aim to detect a conversation group as a whole but do not examine the individual relationships within it. To the best of our knowledge, \cite{liu2022holistic} is the only work that studies both first- and third-person activities from an egocentric point of view; however, it targets the detection of actions that are not related to conversational interactions, and does not generate predictions that are conditioned on specified individuals. Our proposed task is the first to explore exocentric inter-person interactions from egocentric videos.

\vspace{0.1in}

\noindent\textbf{Egocentric Conversation Interaction}.\ 
Reasoning about human social interactions from a first-person perspective has emerged as a prevailing topic in egocentric vision. Earlier works addressed the problem of social behavior classification~\cite{ye2015detecting,ryoo2013first,yonetani2016recognizing}, social saliency prediction \cite{fathi2012social,soo2015social, lai2022eye}, and motion estimation of a conversation partner\cite{yagi2018future,soo2016egocentric,liu20214d}. Our work is most relevant to the Ego4D~\cite{soo2016egocentric} Social Benchmark Talking to Me (TTM) task, which identifies the conversation partners that are talking to the camera wearer. Xue et al.~\cite{xue2023egocentric} introduced a task translation method that achieves state-of-the-art performance on multiple Ego4D benchmark tasks, including TTM. Lin et al.~\cite{lin2023quavf} proposed fusing visual prediction and audio prediction based on face quality score to address the TTM task. 
Jiang et al.~\cite{jiang2022egocentric} introduced the task of Active Speaker Localization (ASL), that entails predicting active speakers in an egocentric scene. Recently, Ryan et al.~\cite{ryan2023egocentric} proposed a Selective Auditory Attention Localization (SAAL) problem, aiming to localize the speaker who is the camera wearer's target of auditory attention. Notably, previous methods are merely a sub-task of our proposed problem of constructing the complete conversational graph. Moreover, we propose the first model that explicitly reasons about the subject-level correlations of the visual and audio signals captured by the egocentric camera to infer their conversational behaviors.


\vspace{0.1in}

\noindent\textbf{Audio-Visual Learning in Egocentric Vision}.\ 
Limited prior work has tackled audio-visual learning in the egocentric setting. Thanks to the recent large-scale datasets~\cite{damen2018scaling,grauman2022ego4d} that contain both visual and audio streams, recent inspiring work integrate cues from both modalities for egocentric action recognition~\cite{kazakos2019epic,xiao2020audiovisual,kazakos2021little,planamente2021cross}, sound object localization in egocentric videos~\cite{huang2023egocentric}, and visual representation learning from audible interactions in egocentric videos~\cite{mittal2022learning}. Another stream of work studies egocentric audio-visual learning in simulated environments, enabling embodied agents to both see and hear in order to perceive 3D environments, and tackles tasks such as embodied navigation~\cite{chen2020soundspaces,chen2020learning, gao2023sonicverse}, echolocation~\cite{gao2020visualechoes}, and scene mapping~\cite{majumder2023chat2map}. Our work proposes a completely new task, wherein we aim to infer exocentric conversational interactions from egocentric videos leveraging both the visual and audio cues. 


\begin{figure}[t]
\centering
\includegraphics[width=1.0\linewidth]{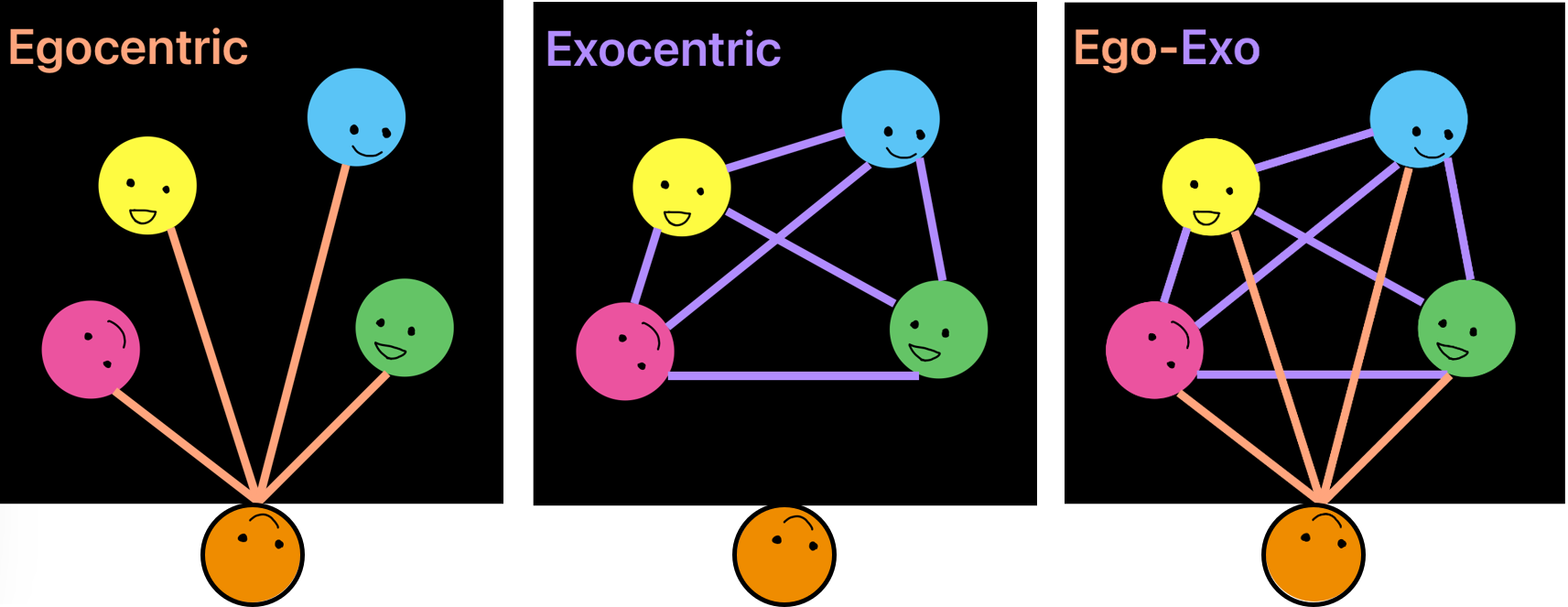}
\caption{\textbf{An illustration of the Conversational Graph.} The left, center, and right figures visualize ${G}_{Ego}$, ${G}_{Exo}$, and ${G}_{Conv}$, respectively. See Sec.~\ref{sec:problem} for details.}
\label{fig:graph}
\end{figure}

\section{The Audio-Visual Conversational Graph}
\label{sec:problem}
We start by formally defining a \emph{Conversational Graph} (Sec. \ref{sec:formulation}), and then introduce the dataset and metrics for the proposed task (Sec. \ref{sec:dataset}).

\subsection{Problem Formulation}\label{sec:formulation}
Given an egocentric video clip $\mathbf{X} \in \mathbb{R}^{C\times T\times H\times W}$ with $T$ indicating the number of frames, our goal is to generate a dynamic directed graph $\mathbf{G}_{Conv}$ that describes the conversational social interactions-- who is looking at and listening to whom, at each moment-- of all social partners in $\mathbf{X}$. Formally, we define the \emph{ Conversational Graph} $\mathbf{G}_{Conv}=(V, E)$, which consists of two connected components $\mathbf{G}_{Ego}$ and $\mathbf{G}_{Exo}$. 

As shown in Fig.~\ref{fig:graph}, $\mathbf{G}_{Ego}=(V_{ego}, E_{ego})$ is a bipartite graph that
describes the conversational interactions between the camera wearer and \textit{each} of the other social partners in the scene, while $\mathbf{G}_{Exo}=(V_{exo}, E_{exo})$ is a non-bipartite graph that describes the interactions among all subjects \textit{except} the camera wearer. 

\begin{figure}[t]
\centering
\includegraphics[width=1.0\linewidth]{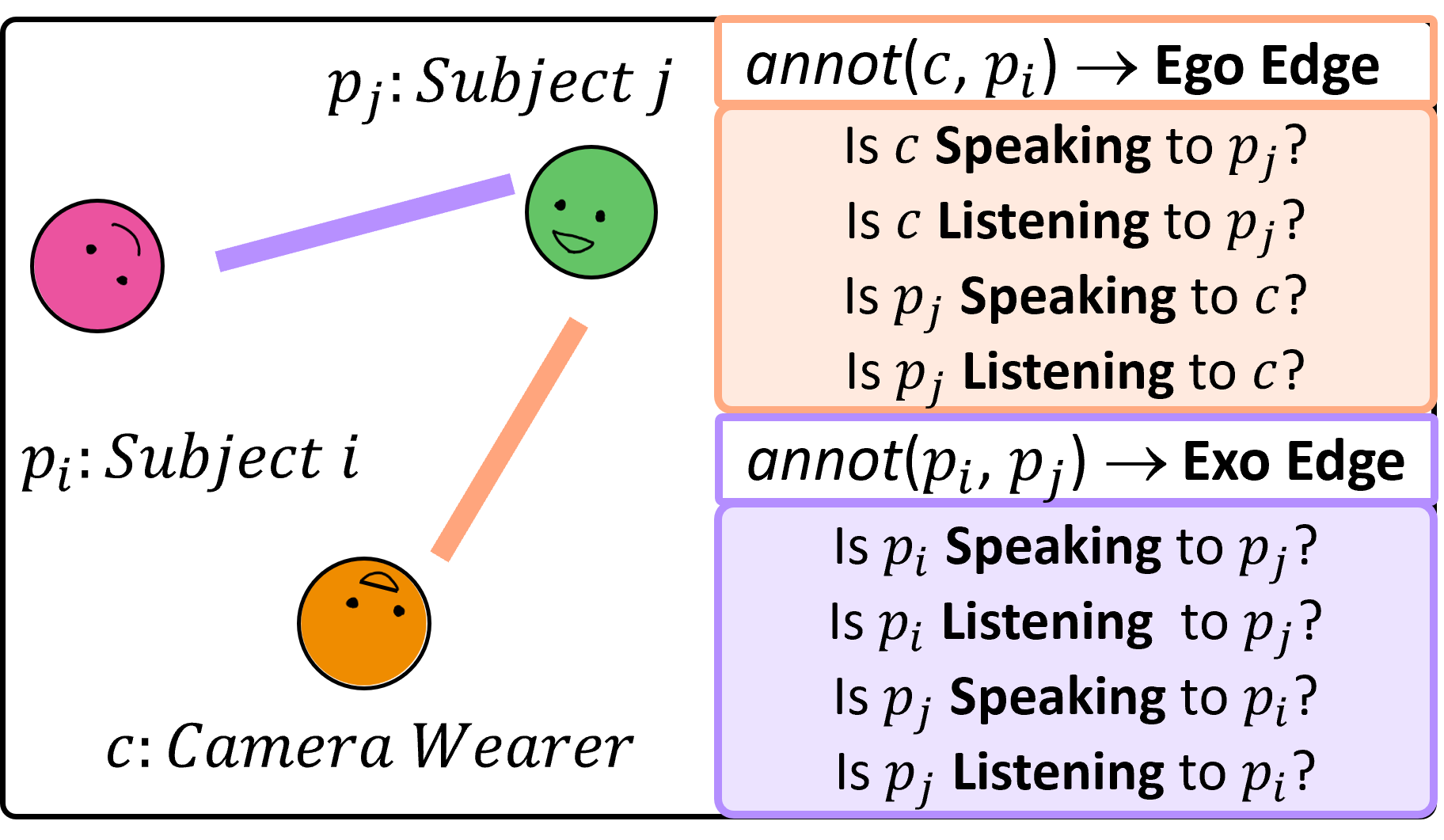}
\vspace{-0.25in}
\caption{\textbf{An example of the edge attributes.} We have binary annotations for each pair of the participants in the conversation, including the camera wear and all other partners. 
}
\label{fig:edge}
\end{figure}

More formally, we define the nodes, edges, and edge attributes of $\mathbf{G}_{Conv}$ as follows: 
\begin{itemize}
    \vspace{0.05in}
    \item \textbf{Nodes:} $V = V_{ego} + V_{exo}$, with:
    \begin{itemize}
    \item $V_{ego} = \{c\}$, where $c$ denotes camera wearer.
    \item $V_{exo} = \{p_1, p_2, \ldots, p_N\}$, where N denotes the number of other social partners besides the camera wearer. 
    \end{itemize}
    
    \vspace{0.05in}
    \item \textbf{Edges:} $E = E_{ego} + E_{exo}$, in which we have:
    \begin{itemize}
    \item $E_{ego} = \{c \rightarrow p_i,\; p_i \rightarrow c \mid {\textrm {for all}}\; p_i\in V_{exo}\}$ 
    \item $E_{exo} = \{p_i \rightarrow p_j,\; p_j \rightarrow p_i \mid {\textrm{for all}}\;p_i,\;p_j\in V_{exo}\;{\textrm {with}}\;i \neq j\;{\textrm {and}}\;i < j\}$
    \end{itemize}
    \vspace{0.05in}
    \item \textbf{Edge Attributes:} For each pair of nodes, we aim to determine (1) whether the two subjects involved are actively engaged in \textbf{Speaking To ($S$)} each other and (2) whether they are actively \textbf{Listening To ($L$)} each other during their social interaction. Therefore, we define the following four types of binary attributes for each pair of nodes:
\[
e^{S}_{c \rightarrow p_i} = 
\begin{cases}
  1 & \text{if $c$ is speaking to $p_i$}\\
  0 & \text{otherwise}
\end{cases}
\]
\[
e^{L}_{c \rightarrow p_i} = 
\begin{cases}
  1 & \text{if $c$ is listening to $p_i$}\\
  0 & \text{otherwise}
\end{cases}
\]
\[
e^{S}_{p_i \rightarrow c} = 
\begin{cases}
  1 & \text{if $p_i$ is speaking to $c$}\\
  0 & \text{otherwise}
\end{cases}
\]
\[
e^{L}_{p_i \rightarrow c} = 
\begin{cases}
  1 & \text{if $p_i$ is listening to $c$}\\
  0 & \text{otherwise}
\end{cases}
\]
\end{itemize}
Similarly, for edges $e_{p_i \rightarrow p_j}$, $e_{p_j \rightarrow p_i}$ in $E_{exo}$, we define the same set of binary attributes for each pair of nodes. An intuitive illustration is shown in Fig.~\ref{fig:edge} for better understanding. These edge attributes fully characterize the conversational interactions among all subjects involved in the scene, and they are independent from each other, because anyone can be speaking to or listening to one another regardless the behaviors of others.

\subsection{Dataset and Annotations}\label{sec:dataset}
\label{sec:data}
The study of directional conversation-related social behaviors has been explored in the \textit{Talking to Me} task from the Ego4D \cite{grauman2022ego4d} Social Benchmarks. However, that task focuses on the social behaviors performed by the other social partners towards the camera wearer, and does not provide annotations between the social partners. Besides, it does not provide any listening-related labels. Therefore, we cannot use it for our proposed task.

Instead, we make use of the recently introduced \textit{Egocentric Concurrent Conversations Dataset} \cite{ryan2023egocentric}, which contains a total of 50 participants, evenly distributed across 10$\sim$30-minute data collection sessions, with each session comprising groups of five individuals. Each individual wears a headset with an Intel SLAM camera and an array of six microphones during the sessions, resulting in $\sim$20 hours of egocentric videos in total. This means in each session with five people in total, each of them serves as the camera wearer in their egocentric video, and the maximum number of the other social partners captured in the egocentric video frames is four. This dataset setting aligns with the prior psychological study \cite{krems2019conversations} which reveals a size constraint of about four people for meaningful group conversations.

To re-purpose the dataset for our problem of computing the audio-visual conversational graph, we generate the ground-truth labels for 4 egocentric interactions, and 6 exocentric interactions in each recorded video leveraging existing annotations. The dataset provides head bounding boxes for all visible faces captured in the egocentric frames, speaking activity annotations of the camera wearer, and the auditory attention target of the camera wearer (as defined and used in their proposed task). Note that these annotations are egocentric-oriented and we need to further synchronize the information collected from different camera wearers within the same data collection session to generate annotations for our exocentric tasks. See the supplementary for details on how we generate the ground-truth annotations.

\section{The Audio-Visual Conversational Attention Model}\label{sec:model}
\begin{figure*}[htbp]
\centering
\includegraphics[width=1.0\linewidth]{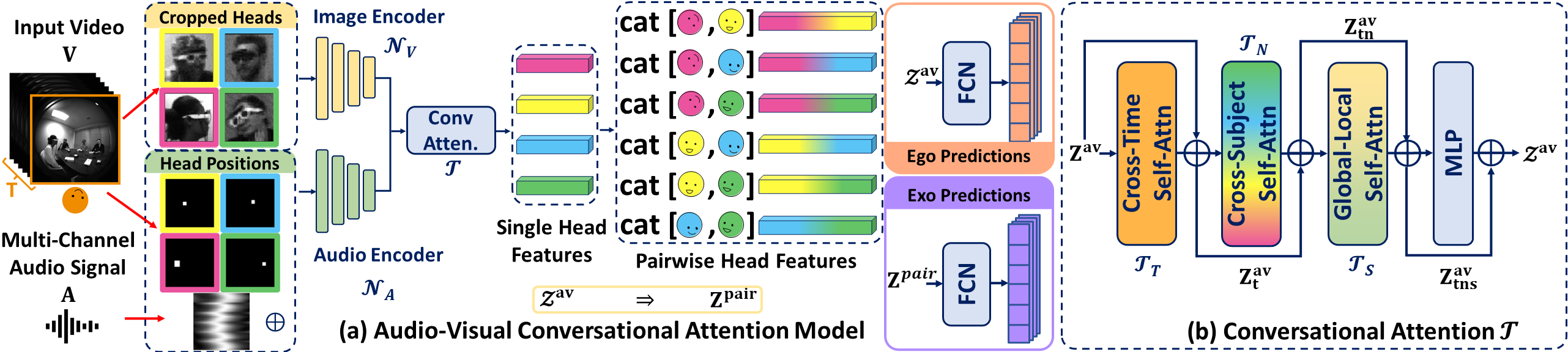}
\vspace{-1.5em}
\captionof{figure}{\textbf{Model Architecture Overview}: Our model takes multiple egocentric frames and multi-channel audio signals. \newline (a) For each frame, the faces of social partners are cropped to serve as raw visual input, while their corresponding head positions are concatenated with audio inputs to generate positional audio signals. Both visual and audio signals are encoded by two separate ResNet18 Backbones and are concatenated to produce Audio-Visual features for each cropped head. (b) After obtaining temporal Audio-Visual feature tubes of video length, they are flattened into a token to be fed into the Conversational Attention Module to produce augmented Single Head Feature feature $\mathcal{Z}^{av}$. Egocentric Classifiers directly take them to predict Egocentric Edge Attributes, and pairs of these features are arbitrarily combined to generate pairwise audio-visual features to predict Exocentric Edge Attributes.}
\vspace{-1.0em}
\label{fig:overview}
\end{figure*}

Predicting the complete conversational interactions of all subjects in the aforementioned conversation graph purely from egocentric videos is nontrivial. The complexity arises from the fact that the conversational dynamics among a social group involve a rich set of signals, including the visual appearance of all social partners, the directional sounds they make, and their spatial locations. Moreover, solving the task requires reasoning of the correlations among these diverse signals across all subjects within the social group. 

In the following, we first explain how we extract audio-visual representations (Sec.~\ref{sec:avfeature}). Then, we present our Audio-Visual Conversational Attention (\name) model that disentangles the correlations of the representations from different subjects with an attention mechanism (Sec.~\ref{sec:attention}). Finally, we introduce the classifiers used to predict the edge attributes (Sec.~\ref{sec:model_classifier}).




\subsection{Audio-Visual Feature Extraction}\label{sec:avfeature}

Below we introduce representations of the location of all speakers, their visual appearance, and multi-channel audio. 

\vspace{0.05in}

\noindent\textbf{Speaker Location}. To efficiently capture the information of each individual, we use the head bounding boxes annotations mentioned in Sec.~\ref{sec:data} to crop all visible heads in the video frames. We also follow ~\cite{ryan2023egocentric} to use binary masks at the locations of the cropped heads, denoted as $\mathbf{S} \in \mathbb{R}^{N \times 1\times T\times h\times w}$, to preserve the spatial positions for all social partners.

\vspace{0.05in}

\noindent\textbf{Head Feature Extraction}. Given an egocentric video clip $\mathbf{V} \in \mathbb{R}^{3\times T\times H\times W}$, a head image tube $\mathbf{H} \in \mathbb{R}^{N \times 3\times T\times h\times w}$ can be generated based on the speaker location information, where $N=4$\footnote{Our model is general in terms of the number of participants $N$, and we use $4$ due to the nature of the dataset we use as described in Sec.~\ref{sec:dataset}.}. The Image Encoder $\mathcal{N}_{V}$ takes the image tube and outputs a downsampled (by a factor of $s$) feature map, followed by a projection layer that reduces the dimension of the feature map. We have $\mathbf{Z}^v \in \mathbb{R}^{N \times C \times T \times \frac{H}{s}\times \frac{W}{s}}$, where $C=256$ is the number of channels. Since each frame may capture a maximum of four individuals, we assume that there are four people present in every frame and use zero paddings to represent those who are not visible in the frame.

\vspace{0.05in}

\noindent\textbf{Audio Feature Extraction}. We follow \cite{jiang2022egocentric, ryan2023egocentric} to concatenate channel correlation features with the real and complex part of the multi-channel spectrogram as our audio input. The edge attributes defined in our task are closely related to the location and the vocal activity of each individual. However, the multi-channel audio input only provides the global information for the entire scene. To obtain the representation of the vocal activity for each social partner in each frame, we sequentially concatenate each person's binary mask $\mathbf{S}$ to the audio input to generate a video-length location-aware multi-channel audio signal $\mathbf{A} \in \mathbb{R}^{N \times C\times T\times H\times W}$ as the input to the Audio Encoder $\mathcal{N}_{A}$. It outputs a downsampled feature map with reduced dimension $\mathbf{Z}^a \in \mathbb{R}^{N \times C \times T \times \frac{H}{s}\times \frac{W}{s}}$, with the same size as $\mathbf{Z}^v$. 

After obtaining the visual feature $\mathbf{Z}^v$ and the audio feature $\mathbf{Z}^a$ for each subject in the frame, we concatenate them along the channel dimension to generate the Audio-Visual feature $\mathbf{Z}^{av} \in \mathbb{R}^{N \times D \times T \times \frac{H}{s}\times \frac{W}{s}}$ with $D=512$ being the number of channels.

\subsection{Conversational Attention}\label{sec:attention}

We propose \textit{Conversational Attention} $\mathcal{T}=\{\mathcal{T}_{T}, \mathcal{T}_{N}, \mathcal{T}_{S}\}$ that applies a self-attention~\cite{vaswani2017attention} mechanism to the extracted audio-visual representations.
This shares a similar spirit to \cite{bertasius2021space}, where they use space-time attention for video understanding. Here we use attention to analyze the conversational interactions across time, subjects, and modalities. We flatten the spatial-temporal dimension of $\mathbf{Z}^{av}$ and obtain ${L=T \times N \times S}$ tokens, with $S=\frac{H}{s}\times \frac{W}{s}$ and an embedding dimension of $D$. We add a linear learnable positional embedding $E\in \mathbb{R}^{L\times D}$ to help preserve positional information for each dimension. 


\vspace{0.05in}

\noindent\textbf{Cross-Time Attention}. Egocentric vision inherently involves drastic scene changes; this makes it crucial to consider information from both current and neighboring frames for a richer temporal context. $\mathcal{T}_{T}$ takes in token $\mathbf{Z}^{av}\in 
\mathbb{R}^{L\times D}$ and applies self-attention to each patch over the temporal dimension. It outputs a token $\mathbf{Z}^{av}_{t}$ that aggregates information across time. 

\vspace{0.05in}

\noindent\textbf{Cross-Subject Attention}. Our task demands a simultaneous prediction of diverse conversational behaviors for all subjects, thus making it essential to allow representations of each subject to attend to each other for understanding the interpersonal dynamics.
$\mathcal{T}_{N}$ processes token $\mathbf{Z}^{av}_{t}$, applying self-attention to tokens at corresponding spatiotemporal locations for each person. This facilitates interactions among audio-visual features of individuals, enhancing prediction distinctions across subjects. It yields token $\mathbf{Z}^{av}_{tn}$.

\vspace{0.05in}

\noindent\textbf{Global-Local Attention}. As depicted in Section~\ref{sec:avfeature}, the head feature is derived from each individual in the frame, and the audio feature contains both scene-wide details and facial positional information. We also apply cross-modality Global-Local self-attention to correlate global audio-positional tokens with local head appearance tokens, enhancing the scene awareness and semantic alignment for each feature. $\mathcal{T}_{S}$ takes in token $\mathbf{Z}^{av}_{tn}$ and compares all patches with each other. 


A residual connection aggregates output from each self-attention layer, and the final feature passes through another fully connected layer. Overall, with a given audio-visual input, we denote the augmented feature $\mathbf{Z}^{av}_{tns}$ as:
\begin{equation}
    \small
    \mathbf{Z}^{av}_{tns} = \mathcal{T}[\mathcal{N}_{V}(V) \bigoplus \mathcal{N}_{A}(A)].
\end{equation}
For simplicity, we use the symbol $\mathcal{Z}^{av}$ to refer to $\mathbf{Z}^{av}_{tns}$ in the following text. 

\subsection{Conversation Edge Attributes Classification}\label{sec:model_classifier}

 We predict the edge attributes for the egocentric graph $G_{Ego}$ directly based on individual-specific feature $\mathcal{Z}^{av}$ for the corresponding subject in the frame. For edges in the exocentric graph $G_{Exo}$, we need to pairwise fuse features for any two subjects of interest to predict the corresponding edge attributes. Next, we introduce how we fuse the pairwise feature for the predictions of the exocentric graph, and our prediction head.

\vspace{0.05in}

\noindent\textbf{Pairwise Feature Fusion}. Since $\mathcal{Z}^{av}$ is individual-specific, we need to perform pairwise fusion to obtain Audio-Visual features for any pair of individuals in the video to predict edge attribute for $G_{Exo}$. Given the maximum possible number of individuals is four in the dataset, we can obtain M ($C_{2}^{4} = 6$) combinations of pairwise features per frame. We directly concatenate the features for the two subjects in each pair to generate the pairwise features $\mathbf{Z}^{pair} \in \mathbb{R}^{M \times P\times T \times \frac{H}{s} \times \frac{W}{s}}$.

\vspace{0.05in}

\noindent\textbf{Prediction Head}. We directly use the refined individual-specific feature $\mathcal{Z}^{av}$ in classifiers for predicting edge attributes of $G_{Ego}$, and use $\mathbf{Z}^{pair}$ for predicting those for $G_{Exo}$. As shown in Section ~\ref{sec:formulation}, we use separate classifiers for each type of binary edge attribute. Therefore, we have a total of eight classifiers to predict the complete set of edge attributes. Each of the classifiers contains a pooling layer and a fully connected layer for predicting logits. We train all eight classifiers together with cross-entropy loss and encourage the model to predict the right attribute and to identify the extra background category.


\begin{table*}[htb]
\footnotesize 
\centering
{
\setlength{\tabcolsep}{8pt} 
\renewcommand{\arraystretch}{1} 
\begin{tabular}{c|c|c|c|c|c|c|c|c|c}
\hline
\multicolumn{1}{c|}{\multirow{2}{*}{Method}}          
&\multicolumn{4}{c|}{Egocentric Graph} &\multicolumn{4}{c}{Exocentric Graph} & \multicolumn{1}{|c}{Ego-Exo Avg} \\ 
\cline{2-10}
\multicolumn{1}{c|}{} & $e^{S}_{c \rightarrow p_i}$ &$e^{S}_{p_i \rightarrow c}$& $e^{L}_{c \rightarrow p_i}$ & $e^{L}_{p_i \rightarrow c}$ &  $e^{S}_{p_i \rightarrow p_j}$ & $e^{S}_{p_j \rightarrow p_i}$ & $e^{L}_{p_i \rightarrow p_j}$ & $e^{L}_{p_j \rightarrow p_i}$ & $G_{conv}$\\ 
\hline \hline
\makecell{SAAL (Acc)}  & 86.23 & 67.10 & 86.48 & 86.99 & / & / & / & / & / \\
\makecell{ASL+Layout (Acc)} & 13.71 & 55.53 & 83.48 & 77.35 & 65.96 & 43.63 & 64.49 & 79.04 & 60.40 \\
\hline
\makecell{\name (Acc)} & \textbf{90.02} & \textbf{75.94} & \textbf{87.80} & \textbf{90.63} & \textbf{86.15} & \textbf{75.89} & \textbf{75.91} & \textbf{85.75} & \textbf{83.51} \\
\hline \hline
\makecell{SAAL} (mAP) & 68.43 & 44.97 & 44.64 & 39.55 &  / & / & / & / & / \\
\makecell{ASL+Layout} (mAP) & \textbf{86.28} & 47.45 & 21.83 & 47.91 & 45.91 & 46.68 & 18.98 & 16.15 & 41.40 \\
\hline

\makecell{\name} (mAP) & 82.08 & \textbf{68.94} & \textbf{60.70} & \textbf{65.48} & \textbf{72.73} & \textbf{63.36} & \textbf{32.35} & \textbf{29.29}  & \textbf{59.37} \\
\hline
\end{tabular}}
\vspace{-0.5em}
\caption{\textbf{Comparing to Prior Work.} There are no pre-exisiting baselines directly comparible for our proposed new task. We devise two baseline methods with components adapted from prior work: SAAL~\cite{ryan2023egocentric} and  ASL~\cite{jiang2022egocentric}+Layout. We report both accuracy and mAP.}
\label{table:baseline}
\end{table*}


\begin{table*}[t]
\footnotesize 
\centering
{
\setlength{\tabcolsep}{8pt} 
\renewcommand{\arraystretch}{1} 
\begin{tabular}{c|c|c|c|c|c|c|c|c|c|c}
\hline
\multicolumn{1}{c|}{\multirow{2}{*}{}}          
&\multicolumn{5}{c|}{Egocentric Graph} &\multicolumn{5}{c}{Exocentric Graph} \\ 
\cline{2-11}
\multicolumn{1}{c|}{} & $e^{S}_{c \rightarrow p_i}$ &$e^{S}_{p_i \rightarrow c}$& $e^{L}_{c \rightarrow p_i}$ & $e^{L}_{p_i \rightarrow c}$ & Ego Avg & $e^{S}_{p_i \rightarrow p_j}$ & $e^{S}_{p_j \rightarrow p_i}$ & $e^{L}_{p_i \rightarrow p_j}$ & $e^{L}_{p_j \rightarrow p_i}$ & Exo Avg \\ 
\hline \hline
\makecell{\textsc{Direct Concat}} & 88.69 & 68.65 & 83.83 & 86.85 & 82.00 & 67.60 & 69.53 & 85.07 & 83.52 & 76.43 \\
\hline
\makecell{\name (T)} & 89.49 & 73.60 & 86.88 & 87.43 & 84.35 & 72.97 & 74.57 & 85.13 & 84.36 & 79.26 \\
\makecell{\name (N)} & 88.62 & 68.83 & 85.12 & 88.30 & 82.72 & 68.39 & 69.34 & 85.72 & 84.36 & 76.95 \\
\makecell{\name (S)} & 88.62 & 69.38 & 85.11 & 87.47 & 82.65 & 68.11 & 70.23 & 85.09 & 83.96 & 76.85 \\
\hline
\makecell{\name (TN)} & 89.58 & 75.04 & 87.05 & 88.89 & 85.14 & 74.59 & 75.12 & 85.91 & 85.44 & 80.27 \\
\makecell{\name (TS)} & 89.36 & 75.12 & 86.57 & 88.29 & 84.84 & 75.42 & 74.76 & 85.23 & 84.25 & 79.92 \\
\makecell{\name (NS)} & 89.73 & 74.23 & 87.52 & 88.81 & 85.07 & 73.60 & 74.37 & 86.53 & 85.51 & 80.00 \\
\hline \hline
\makecell{\name} & \textbf{90.02} & \textbf{75.94} & \textbf{87.80} & \textbf{90.63} & \textbf{86.15} & \textbf{75.89} & \textbf{75.91} & \textbf{86.61} & \textbf{85.75} & \textbf{81.04} \\

\hline
\end{tabular}}
\vspace{-0.5em}
\caption{\textbf{Ablation Study on Conversational Attention}. To explore the separate impact of cross-time attention, cross-subject attention, and Gloabl-Local attention, we exhaustively explore different combinations of the components in our conversational attention model. We report the classification accuracy and see Supp. for the mAP results.
}
\label{table:modeldesign}
\end{table*}

\begin{table*}[t]
\footnotesize 
\centering
{
\setlength{\tabcolsep}{8pt} 
\renewcommand{\arraystretch}{1} 
\begin{tabular}{c|c|c|c|c|c|c|c|c|c|c}
\hline
\multicolumn{1}{c|}{\multirow{2}{*}{}}          
&\multicolumn{5}{c|}{Egocentric Graph} &\multicolumn{5}{c}{Exocentric Graph} \\ 
\cline{2-11}
\multicolumn{1}{c|}{} & $e^{S}_{c \rightarrow p_i}$ &$e^{S}_{p_i \rightarrow c}$& $e^{L}_{c \rightarrow p_i}$ & $e^{L}_{p_i \rightarrow c}$ & Ego Avg & $e^{S}_{p_i \rightarrow p_j}$ & $e^{S}_{p_j \rightarrow p_i}$ & $e^{L}_{p_i \rightarrow p_j}$ & $e^{L}_{p_j \rightarrow p_i}$ & Exo Avg \\ 
\hline \hline
\makecell{\textsc{Head Only}} & 63.18 & 57.76 & 79.34 & 80.39 & 70.17 & 57.51 & 58.43 & 84.58 & 82.98 & 70.81 \\
\makecell{\textsc{Audio Only}} & 88.57 & 59.34 & 77.12 & 76.97 & 75.50 & 32.47 & 33.60 & 67.43 & 71.84 & 51.34 \\
\makecell{\textsc{Mask Only}} & 63.33 & 58.32 & 81.03 & 80.21 & 70.72 & 57.57 & 58.75 & 84.96 & 84.09 & 71.19 \\
\hline
\makecell{\textsc{Head+Mask}} & 64.45 & 59.18 & 80.86 & 80.92 & 71.31 & 59.66 & 59.17 & 84.50 & 81.59 & 71.23 \\
\makecell{\textsc{Audio+Mask}} & 89.20 & 75.29 & \textbf{87.89} & 87.74 & 85.03 & 75.67 & 74.74 & 84.64 & 84.06 & 79.78 \\
\hline \hline
\makecell{\name} & \textbf{90.02} & \textbf{75.94} & 87.80 & \textbf{90.63} & \textbf{86.15} & \textbf{75.89} & \textbf{75.91} & \textbf{86.61} & \textbf{85.75} & \textbf{81.04} \\
\hline
\end{tabular}}
\vspace{-0.5em}
\caption{\textbf{Modality Ablation}. As described in ~\ref{sec:avfeature}, our input signals consist of three components: head images, multi-channel audio, and binary position masks. In this ablation study, 
we explore different choices of input signals, and assess their relative impact on the final performance. We report the classification accuracy and see Supp. for the mAP results.
}
\label{table:ablation-modality}
\end{table*}



\section{Experiments}
We validate our approach by comparing it with a series of baselines, and present both quantitative comparisons and qualitative visualizations.

\subsection{Implementation Details}
Both the visual and audio encoder are adapted from the ResNet-18 backbone pre-trained on ImageNet-1K, which is composed of four convolutional blocks. The \name model is composed of two 8-head self-attention blocks \cite{vaswani2017attention}, and we use similar implementations as in \cite{bertasius2021space}. The model is implemented in PyTorch, and trained for around 9 epochs using an Adam optimizer with a learning rate of 1e-4. 

\subsection{Baselines}
Since we are the first to address the challenging problem of conversational graph,there are no pre-existing baselines directly comparable. We adapt two baselines from prior work~\cite{ryan2023egocentric,jiang2022egocentric}, and various ablated versions of our method. 

\begin{itemize}
    \item \textbf{SAAL}~\cite{ryan2023egocentric}: SAAL was originally designed to predict the camera wearer's auditory attention from an egocentric point of view, corresponding to $e^{L}_{c \rightarrow p_i}$ in our problem setting. To extend its applicability to our broader egocentric-related tasks, we adapt our annotations to their setting and add extra prediction layers to its decoder for the other egocentric tasks we tackle. 
    \vspace{0.05in}
    \item \textbf{ASL~\cite{jiang2022egocentric} + Layout}: This is a heuristics baseline that combines 3D person layout estimation and the active speaker localization (ASL) to infer the ego and exo conversational interactions. Using single view depth estimation~\cite{ranftl2020towards} and a 3D head pose regression model, we predict the participants' 3D head locations and facing directions in the camera wearer's frame. We consider the interaction probability of two subjects proportional to the angle between the vector from one person to another and the the facing direction of the person. We multiply the interaction probability and the voice activity probability to generate the final predictions for each edge attribute.
    \vspace{0.05in}
    \item \textbf{\textsc{Direct Concat}}: We exclude the entire Conversational Attention $\mathcal{T}$ component for this baseline. The feature representation $\mathbf{Z}^{av}$ is obtained by directly concatenating $\mathbf{Z}^{v}$ and $\mathbf{Z}^{a}$ without additional augmentation.
    \vspace{0.05in}
    \item \textbf{\name (T, N, S)}: To explore the separate impact of Cross-Time attention (T), Cross-Subject attention (N), and Global-Local attention (S), we use only $\mathcal{T}_{T}$, $\mathcal{T}_{N}$, and $\mathcal{T}_{S}$, respectively from our model for feature fusion while keeping all other settings the same.
    \vspace{0.05in}
     \item \textbf{\name (TN, TS, NS)}: This is the same as the previous three baselines except that we use two types of attention for aggregating the features. In particular, \name (TS) uses similar spatiotemporal attention mechanism commonly employed in various video understanding work~\cite{ryan2023egocentric, bertasius2021space}.
    \vspace{0.05in}
     \item \textbf{\textsc{Head/Audio/Mask Only}}: To explore the impact of different input modalities, we retain only the head image input (Head), the multi-channel audio signal (Audio), or the positional binary masks (Mask) for predicting the edge attributes of the conversational graph.
    \vspace{0.05in}
     \item \textbf{\textsc{Head+Mask / Audio+Mask}}: This is the same the previous three baselines except that we discard either the head image or the audio input. 
\end{itemize}





\subsection{Quantitative Results}
\label{sec:results}

Table~\ref{table:baseline}, \ref{table:modeldesign}, \ref{table:ablation-modality} show our results for comparing with the prior methods, ablation study on our conversational attention design, ablation study on input modalities, respectively. In all three tables, the best results are highlighted with \textbf{boldface}. Since the prediction of each edge attribute is a binary classification problem. We report the accuracy and mean Average Precision (mAP) for each edge attribute classification. Additionally, we present the average performance for egocentric and egocentric tasks. Namely, $e^{S}_{c \rightarrow p_i}$, $e^{S}_{p_i \rightarrow c}$, $e^{L}_{c \rightarrow p_i}$, $e^{L}_{p_i \rightarrow c}$ of $\mathbf{G}_{Ego}$ represent the conversational interaction between the camera wear and other social partner, while $e^{S}_{p_j \rightarrow p_i}$, $e^{S}_{p_i \rightarrow p_j}$, $e^{L}_{p_j \rightarrow p_i}$, $e^{L}_{p_i \rightarrow p_j}$ of $\mathbf{G}_{Exo}$ represent the pair-wise conversational interaction between arbitrary two social partners.




\vspace{0.05in}

\noindent \textbf{Comparing to Prior Work.} 
As shown in Table~\ref{table:baseline}, our \name model consistently outperforms both the SAAL and ASL+Layout baselines across all sub-tasks. For SAAL, we outperform it by an average of approximately $~4.45\%$ on egocentric-related tasks ($+3.99\%$, $+8.84\%$, $+1.32\%$, $+3.64\%$, respectively), demonstrating the benefit of learning these closely related sub-tasks jointly. We observe that the sub-task with the smallest margin is $e^{L}_{c \rightarrow p_i}$, and this is because SAAL is tailored to address this sub-task, thus making the model design and the fine-tuned model parameters well suited for this task. Compared with the heuristic baseline ASL+Layout, our model outperform it by a large margin for both the egocentric and exocentric edges. This further demonstrates that this is a challenging task, which can not be easily solved just by leveraging the people layout and the results from active speaker localization.

\vspace{0.05in}

\noindent \textbf{Ablation Study on Conversational Attention.} In Table~\ref{table:modeldesign}, we investigate how each component and their combinations in the Conversational Attention module $\mathcal{T}$ contribute to the overall performance. 
We can see that \textsc{Direct Concat} leads to the worst performance across almost all tasks, particularly on identifying the exocentric speakers. Using either the Cross-Time attention, Cross-Subject attention, or the Gloabl-Local attention all positively contribute to the final performance. Noticeably, attention across time leads to the largest gain, suggesting the importance of aggregating information from nearby frames to more reliably detect speech activities.

Our final model leverages all three types of attention mechanisms to build the conversational attention block, resulting in a more comprehensive understanding of conversational interactions in egocentric videos. It outperforms the \textsc{Direct Concat} approach by an average of approximately $~4.15\%$ on egocentric-related tasks ($+1.33\%$, $+7.29\%$, $+3.97\%$, $+3.78\%$, respectively) and $~4.61\%$ on exocentric-related tasks ($+8.29\%$, $+6.38\%$, $+1.54\%$, $+2.23\%$, respectively). 
The Cross-Time attention enables the model to capture temporal dependencies and aggregate information from adjacent frames, enhancing its ability to detect voice activities over time. The Cross-Subject attention
contributes by effectively comparing and distinguishing features from different individuals, particularly those related to social partners' relative positions in the frame. The Global-Local attention focuses on features extracted from the appearance of input signals, capturing orientations of heads, movement of lips, and facial expressions. The model achieves a synergistic effect, benefiting from the complementary nature of each attention type.  

\begin{figure*}[t]
\centering
\includegraphics[width=1.0\linewidth]{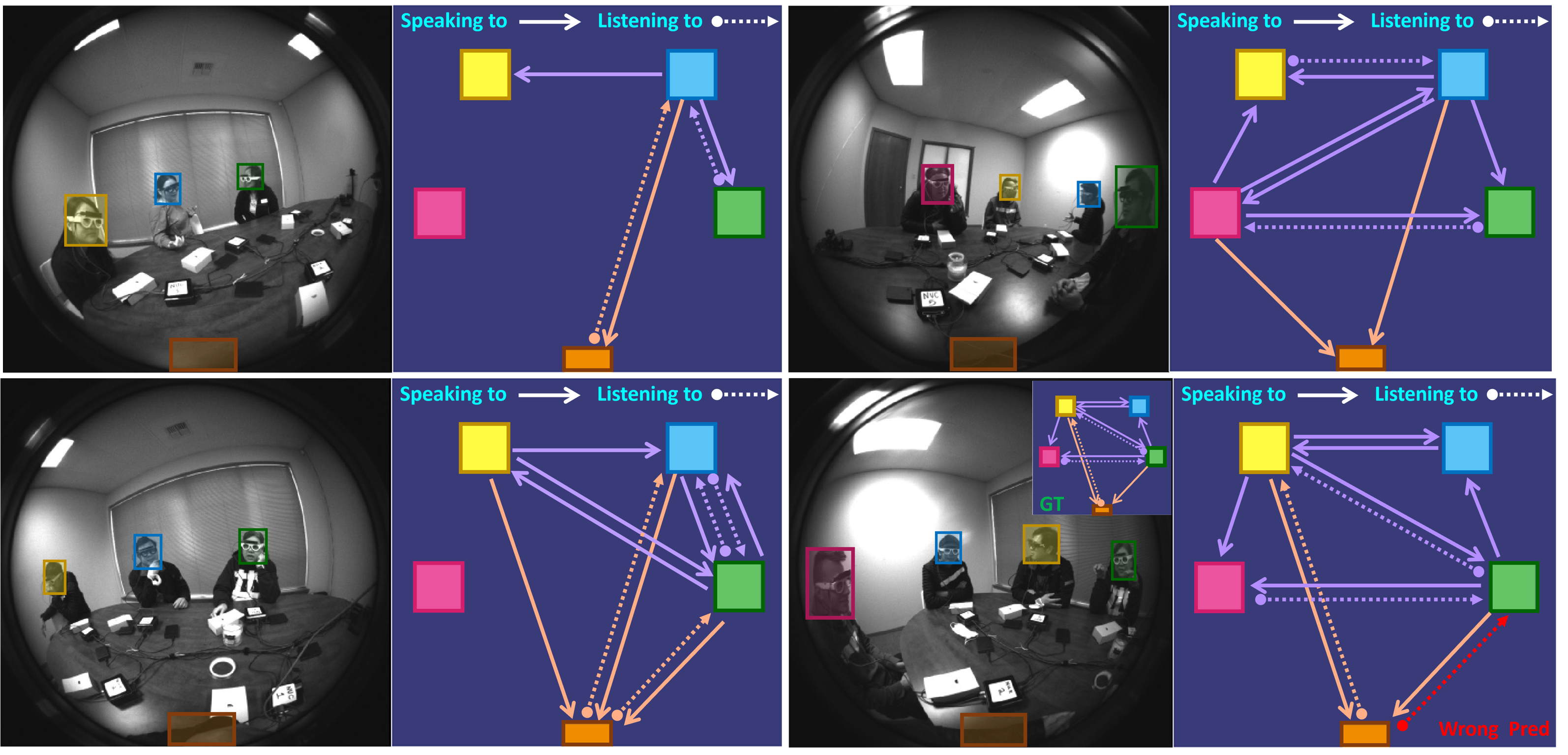}
\vspace{-1.5em}
\captionof{figure}{\textbf{Visualization of the Ego-Exocentric Conversational Graph from our model prediction.} We show three successful cases and one failure case in the bottom right. For the last failure example, we also overlay the ground truth of the conversational graph on the top right corner of the video frame as reference.}

\label{fig:viz}
\end{figure*}

\vspace{0.05in}

\noindent \textbf{Ablation Study on Input Modality.} As illustrated in Sec.~\ref{sec:avfeature}, our input modalities consist of three components: 1) \textit{Heads images} cropped from the egocentric frames, which are subject-specified. 2) \textit{Audio input} containing global cross-correlation features of multi-channel audio from the egocentric video. 3) \textit{Positional Binary Mask}, specifying each individual's location in the frame and serving as an intermediate global-local, subject-specified representation. 

In Table~\ref{table:ablation-modality}, we show the results of evaluating the model's performance by selectively excluding one or more modalities to discern their contributions. As expected, using only the head images leads to significant performance degradation on all speaking-related tasks, as the reasoning of the speaking behaviors needs audio. On the other hand, using only audio produces good results on $Ego_{S}$ while failing on the other tasks, implying the necessity of local features that relate to conversational partners. Using only the positional mask has similar performance as using only the head image, but it infers the potential relationships among social partners based on the abstract representation of their head locations in the entire scene. Interestingly, it outperforms \textsc{Head Only} on listening sub-tasks, probably because they rely more on the position information of people in the space.  

When omitting either the audio signal (\textsc{Head+Mask}) or the visual signal (\textsc{Audio+Mask}), we find that the multi-channel audio is more useful when combined with the positional mask, which also agrees with findings in prior work~\cite{ryan2023egocentric}. This audio-visual modality input provides both global audio activity information and a local abstracted visual prior of all social partners' positions from an egocentric point of view, including interpersonal relative location relationships. It outperforms \textsc{Heads+Mask} by an average of approximately $~13.72\%$ on egocentric-related tasks ($+24.75\%$, $+16.11\%$, $+7.03\%$, $+6.82\%$, respectively) and $~8.55\%$ on exocentric-related tasks ($+16.01\%$, $+15.57\%$, $+0.14\%$, $+2.47\%$, respectively). The main contribution comes from speaking-related tasks. When adding all three modalities as our input, we achieve the best performance.

The above results underscores the synergistic contribution of each modality to the overall effectiveness of the proposed model in predicting egocentric social interactions. The combination of subject-specified visual cues, audio information, and spatial context through the positional mask prove to be essential for achieving comprehensive and accurate predictions.

\subsection{Visualization}
\label{sec:visualization}
In Fig.~\ref{fig:viz}, we visualize some examples from the \textit{Egocentric Concurrent Conversations Dataset} and show the prediction results of the conversational graph from our model. We can see that it is a very challenging problem, as the subjects in the video frames are all visually similar and exhibit complex conversational interactions---there can be multiple people speaking and listening at the same time. Despite the challenges and diverse people layout, our model makes accurate predictions of the complete conversation behaviors for both the camera wearer as well as all other social partners present in the scene from just the egocentric videos, demonstrating the effectiveness of our \name model.

The bottom right corner shows a typical failure case, where our model mistakenly predicts that the camera wearer is listening to the subject in green. We suspect it's because the subjects in blue and green are both speaking at the same time and they are physically very close to each other, so that the model finds hard to tell who is speaking. It would be interesting future work to explore better audio representation to capture more high-resolution spatial information from the multi-channel audio.


\section{Conclusion}


We presented the Audio-Visual Conversational Graph Prediction task to infer exocentric conversational interactions from egocentric videos. Our \name model shows superiority on a challenging multi-speaker, multi-conversation egocentric dataset, paving the way for learning exocentric social interactions from egocentric videos. Future plans involve extending our framework to tackle additional social behavior tasks like gaze prediction and exploring more intricate social relationships within conversation groups. 



\newpage


{
    \small
    \bibliographystyle{ieeenat_fullname}

    \bibliography{main.bbl}
}
\newpage



\setcounter{section}{0}
The supplementary materials consist of: 
\begin{enumerate}[label={}]
    \item \hyperref[sec:anno]{1.} Annotation Details.
    \item \hyperref[sec:modalarch]{2.} \name Model Architectures. 
    \item \hyperref[sec:cost]{3.} Computational Cost and Scalability. 
    \item \hyperref[sec:mapabla]{4.} Ablation Studies with mAP Results.
    \item \hyperref[sec:quali]{5.} More \name Qualitative Results.
    \item \hyperref[sec:limit]{6.} Limitation and Future Work. 
    \item \hyperref[sec:video]{7.} Demo Video. 
\end{enumerate}
\section{Annotation Details}
\label{sec:anno}
Here we introduce the details on how we obtain the labels for the edge attributes of the conversational graph in our task. Ideally, we should annotate each individual's intention of speaking and listening behavior per moment. However, such densely annotated per-moment intention labels are not available in the \textit{Egocentric Concurrent Conversations Dataset} we use, and also hard to obtain in practice. The \textit{Egocentric Concurrent Conversations Dataset} pre-defines two two- or three-people conversational groups per session, and all five participants are instructed to engage in conversations with their own group. In this way, each participant selectively listens to others who belong to their same group in their sessions with multiple self-driven concurrent speakers, allowing for close estimation of auditory attention ground truth. They are synchronized between all participants' annotations to construct the \textbf{Listening To} ground-truth labels in edge attributes. Unlike the \textbf{Listening To} label, we define the \textbf{Speaking To} behavior as not selective, as speaking is a spontaneous behavior and the intentions of speakers are covert hence difficult to quantify. Similarly, they are spread into each ego- and exocentric edges to represent the speaking attributes in each conversational edge. 

A statistical analysis of annotations in the dataset we used reveals that the ratio of positive to negative “Listening To” labels is approximately 1:2, while the ratio of “Speaking To” labels is roughly 1:1. A \textsc{Random Guess} baseline on a subset achieves accuracies of $24.17\%$ and $53.75\%$ for Egocentric and Exocentric Average Performances, respectively.



\section{\name Model Architectures}
\label{sec:modalarch}
\begin{figure}[t]
\centering
\includegraphics[width=1.0\linewidth]{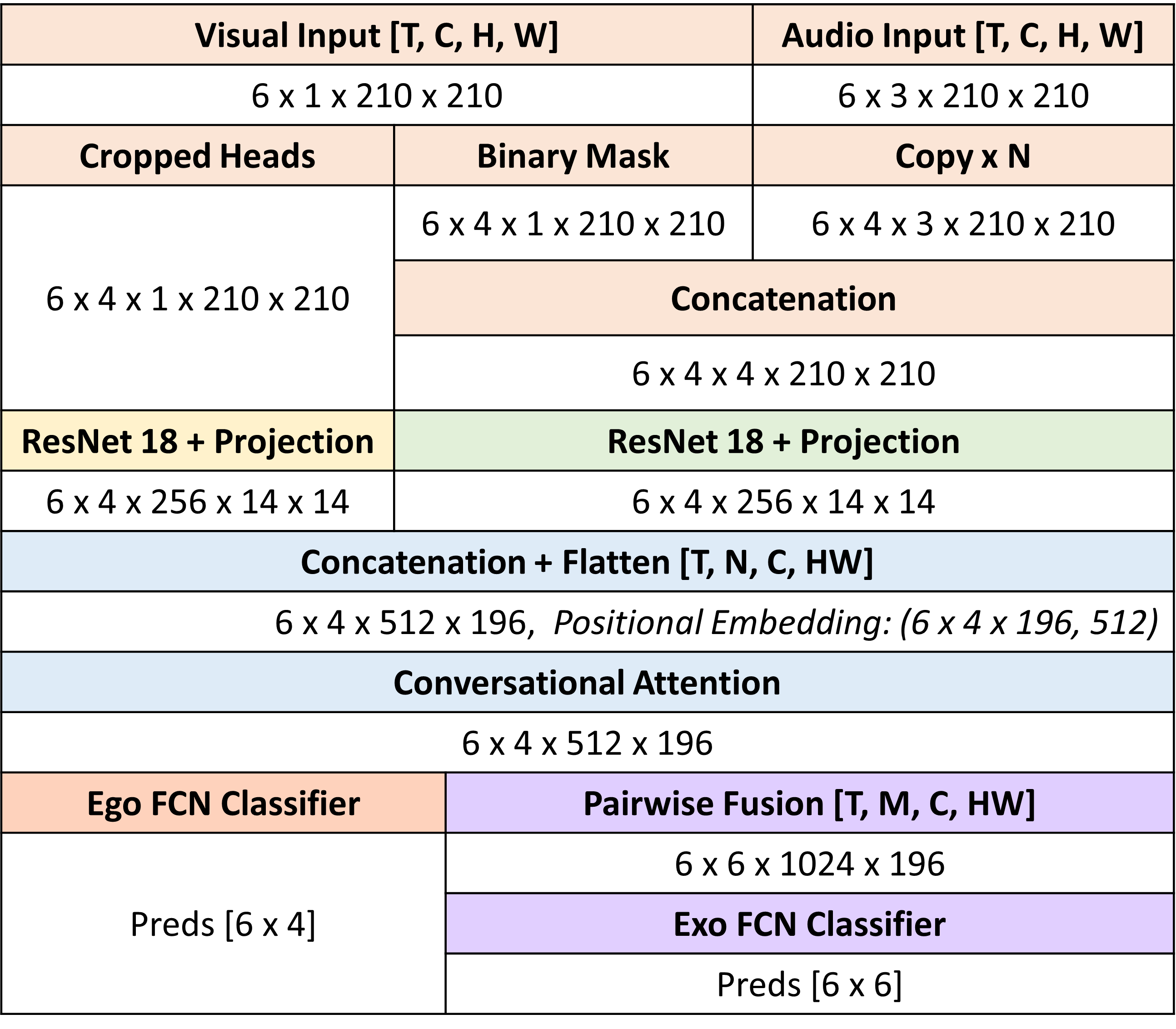}
\vspace{-1.5em}
\captionof{figure}{Architecture Details of \name}

\label{fig:avconv-arch}
\end{figure}

The input egocentric frames $\mathbf{V}$, multi-channel audio signals $\mathbf{A}$, and binary mask $\mathbf{S}$ are all resized to 210$\times$210 for proper alignment. To capture the evolution of the conversational graph through a longer temporal stride setting, each instance input in our experiments consists of 6 frames with a temporal stride of 15, spanning a 90-frame window equivalent to 3 seconds. This results in 15682/6329 (Train/Val) audio-visual samples. Predictions are made all at once on each frame, corresponding to a 0.5-second interval. 
We provide model details for \name in Fig.~\ref{fig:avconv-arch}.

\section{Computational Cost and Scalability}
\label{sec:cost}
\name costs 53M parameters and 25.28 GFLOPs, and can generalize to different numbers of faces such as larger groups with more visible heads, though with an increased computational cost. For example, additional analysis shows that our model uses the same amount of GFLOPs if there are six people present in the scene, and we can still train our model on 2 GeForce RTX 4090s with a batch size of 4. It is because in our setting, the nature of handling more faces simply equals enlarging the batch size, thus not resulting in more operations.

\section{Ablation Studies with mAP Results}
\label{sec:mapabla}
\subsection{Ablation Study on Conversational Attention}
With the second metric mAP, we observe a similar pattern in Table~\ref{table:modeldesign-map} as it in the main paper. Our final model \name outperforms the \textsc{Direct Concat} baseline by an average of $7.18\%$ on almost all egocentric-related tasks ($-2.48\%$, $+6.98\%$, $+14.28\%$, $+11.27\%$) and $9.35\%$ on all exocentric-related tasks ($+7.80\%$, $+5.17\%$, $+15.09\%$, $+10.98\%$). 

\subsection{Ablation Study on Input Modality}
However, patterns in Table~\ref{table:ablation-modality-map} are slightly different from those in main paper. While \textsc{Mask Only} still marks the best performance among almost all single-modality ablations, omitting either the audio signal (\textsc{Head+Mask}) or the visual signal (\textsc{Audio+Mask}) results in a drop in recall on all tasks. This suggests that both audio and visual signals play an important role in the model's performance, and that combining them through multi-modal fusion is crucial for achieving optimal results. Results with using all modality outperforms \textsc{Mask Only} by an average of $24.41\%$ on all egocentric-related tasks ($+27.53\%$, $+16.76\%$, $+21.43\%$, $+28.19\%$) and $16.1\%$ on all exocentric-related tasks ($+17.73\%$, $+16.07\%$, $+17.42\%$, $+13.2\%$). 


\begin{table*}[t]
\footnotesize 
\centering
{
\setlength{\tabcolsep}{8pt} 
\renewcommand{\arraystretch}{1} 
\begin{tabular}{c|c|c|c|c|c|c|c|c|c|c}
\hline
\multicolumn{1}{c|}{\multirow{2}{*}{}}          
&\multicolumn{5}{c|}{Egocentric Graph} &\multicolumn{5}{c}{Exocentric Graph} \\ 
\cline{2-11}
\multicolumn{1}{c|}{} & $e^{S}_{c \rightarrow p_i}$ &$e^{S}_{p_i \rightarrow c}$& $e^{L}_{c \rightarrow p_i}$ & $e^{L}_{p_i \rightarrow c}$ & Ego Avg & $e^{S}_{p_i \rightarrow p_j}$ & $e^{S}_{p_j \rightarrow p_i}$ & $e^{L}_{p_i \rightarrow p_j}$ & $e^{L}_{p_j \rightarrow p_i}$ & Exo Avg \\ 
\hline \hline
\makecell{\textsc{Direct Concat}} & 84.56 & 61.96 & 46.42 & 54.21 & 62.12 & 64.93 & 58.19 & 17.26 & 18.31 & 40.08 \\
\hline
\makecell{\name (T)} & 83.75 & 67.78 & 56.15 & 57.83 & 66.38 & 70.41 & \textbf{64.50} & 21.89 & 22.06 & 44.71 \\
\makecell{\name (N)} & 83.24 & 63.59 & 52.73 & 57.35 & 64.23 & 66.46 & 59.25 & 23.50 & 25.11 & 43.58 \\
\makecell{\name (S)} & 83.85 & 61.16 & 47.48 & 54.42 & 61.73 & 64.41 & 56.72 & 20.05 & 20.47 & 40.41 \\
\hline
\makecell{\name (TN)} & 84.71 & 66.80 & 55.12 & 56.65 & 65.80 & 68.78 & 63.51 & 24.83 & \textbf{30.21} & 46.83 \\
\makecell{\name (TS)} & \textbf{84.92} & 67.04 & 54.02 & 58.46 & 66.11 & 70.00 & 63.32 & 21.64 & 23.42 & 44.60 \\
\makecell{\name (NS)} & 84.41 & 63.49 & 53.22 & 55.10 & 64.05 & 66.13 & 59.31 & 22.05 & 22.45 & 42.49 \\
\hline\hline
\makecell{\name} & 82.08 & \textbf{68.94} & \textbf{60.70} & \textbf{65.48} & \textbf{69.30} & \textbf{72.73} & 63.36 & \textbf{32.35} & 29.29 & \textbf{49.43} \\
\hline
\end{tabular}}
\vspace{-0.5em}
\caption{\textbf{Ablation Study on Conversational Attention in mAP}. As described in the main paper, we report the classification mAP results.}
\label{table:modeldesign-map}
\end{table*}

\begin{table*}[t]
\footnotesize 
\centering
{
\setlength{\tabcolsep}{8pt} 
\renewcommand{\arraystretch}{1} 
\begin{tabular}{c|c|c|c|c|c|c|c|c|c|c}
\hline
\multicolumn{1}{c|}{\multirow{2}{*}{}}          
&\multicolumn{5}{c|}{Egocentric Graph} &\multicolumn{5}{c}{Exocentric Graph} \\ 
\cline{2-11}
\multicolumn{1}{c|}{} & $e^{S}_{c \rightarrow p_i}$ &$e^{S}_{p_i \rightarrow c}$& $e^{L}_{c \rightarrow p_i}$ & $e^{L}_{p_i \rightarrow c}$ & Ego Avg & $e^{S}_{p_i \rightarrow p_j}$ & $e^{S}_{p_j \rightarrow p_i}$ & $e^{L}_{p_i \rightarrow p_j}$ & $e^{L}_{p_j \rightarrow p_i}$ & Exo Avg \\ 
\hline\hline
\makecell{\textsc{Head Only}} & 51.20 & 51.65 & 37.19 & 29.38 & 42.36 & 54.52 & 48.12 & 16.48 & 17.33 & 34.11 \\
\makecell{\textsc{Audio Only}} & \textbf{84.32} & 53.43 & 22.94 & 24.26 & 46.24 & 51.63 & 43.89 & 14.17 & 15.58 & 31.32 \\
\makecell{\textsc{Mask Only}} & 54.55 & 52.18 & 39.27 & 33.54 & 44.89 & 55.00 & 47.29 & 14.93 & 16.09 & 33.33 \\
\hline
\makecell{\textsc{Head+Mask}} & 47.84 & 50.28 & 35.80 & 22.38 & 39.08 & 52.85 & 45.90 & 14.83 & 15.89 & 32.37 \\
\makecell{\textsc{Audio+Mask}} & 45.83 & 47.40 & 22.83 & 21.31 & 34.34 & 50.40 & 43.86 & 14.76 & 15.95 & 31.24 \\
\hline \hline
\makecell{\name} & 82.08 & \textbf{68.94} & \textbf{60.70} & \textbf{65.48} & \textbf{69.30} & \textbf{72.73} & \textbf{63.36} & \textbf{32.35} & \textbf{29.29} & \textbf{49.43} \\
\hline
\end{tabular}}
\vspace{-0.5em}
\caption{\textbf{Modality Ablation in mAP}. As described in the main paper, we report the classification mAP results.}
\label{table:ablation-modality-map}
\end{table*}

\section{More Qualitative Results}
\label{sec:quali}
\begin{figure*}[t]
\centering
\includegraphics[width=1.0\linewidth]{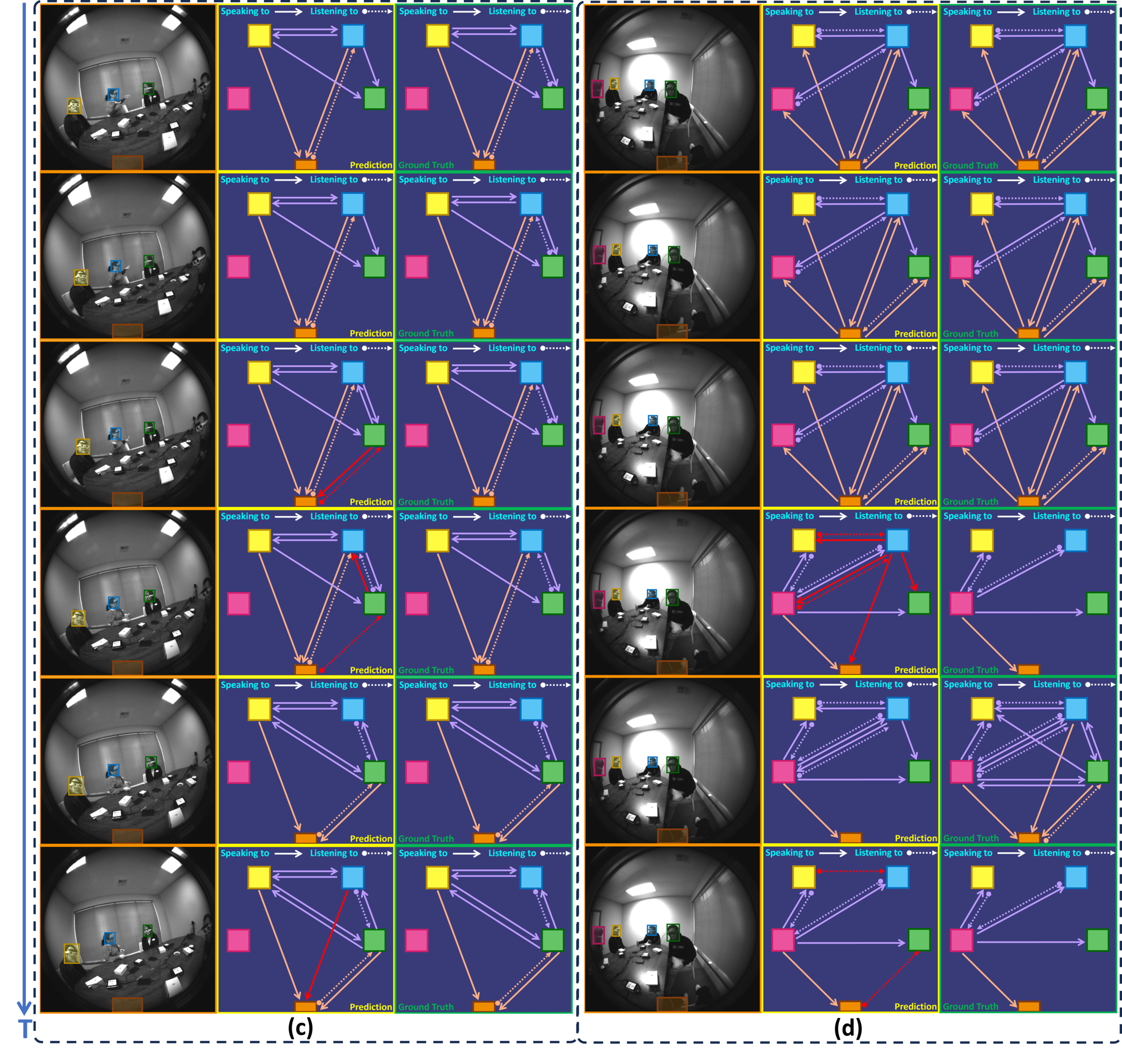}
\captionof{figure}{Visualization of the Ego-Exocentric Conversational Graph from our model prediction.}
\label{fig:suppl_viz_2}
\end{figure*}

In Fig.~\ref{fig:suppl_viz_2}, we provide four additional qualitative results of our model's predictions to further illustrate its performance. In each column, all 6 visualizations come from consecutive input frames of the same validation instance that spans 3 seconds. For each visualization, we present the raw visual input, predictions from our model, and the ground-truth $G_{Conv}$. For each sequence, the ground-truth $G_{Conv}$ changes 1-3 times, resulting in 2-4 evolutions of the conversational graph. Our prediction is able to accurately capture this very challenging graph evolution behavior. When the graph suddenly changes with a drastic difference (in Fig.~\ref{fig:suppl_viz_2}(d)), our model fails to capture all changes but is still able to capture most of them while producing some wrong guesses.

\begin{figure*}[t]
\centering
\includegraphics[width=1.0\linewidth]{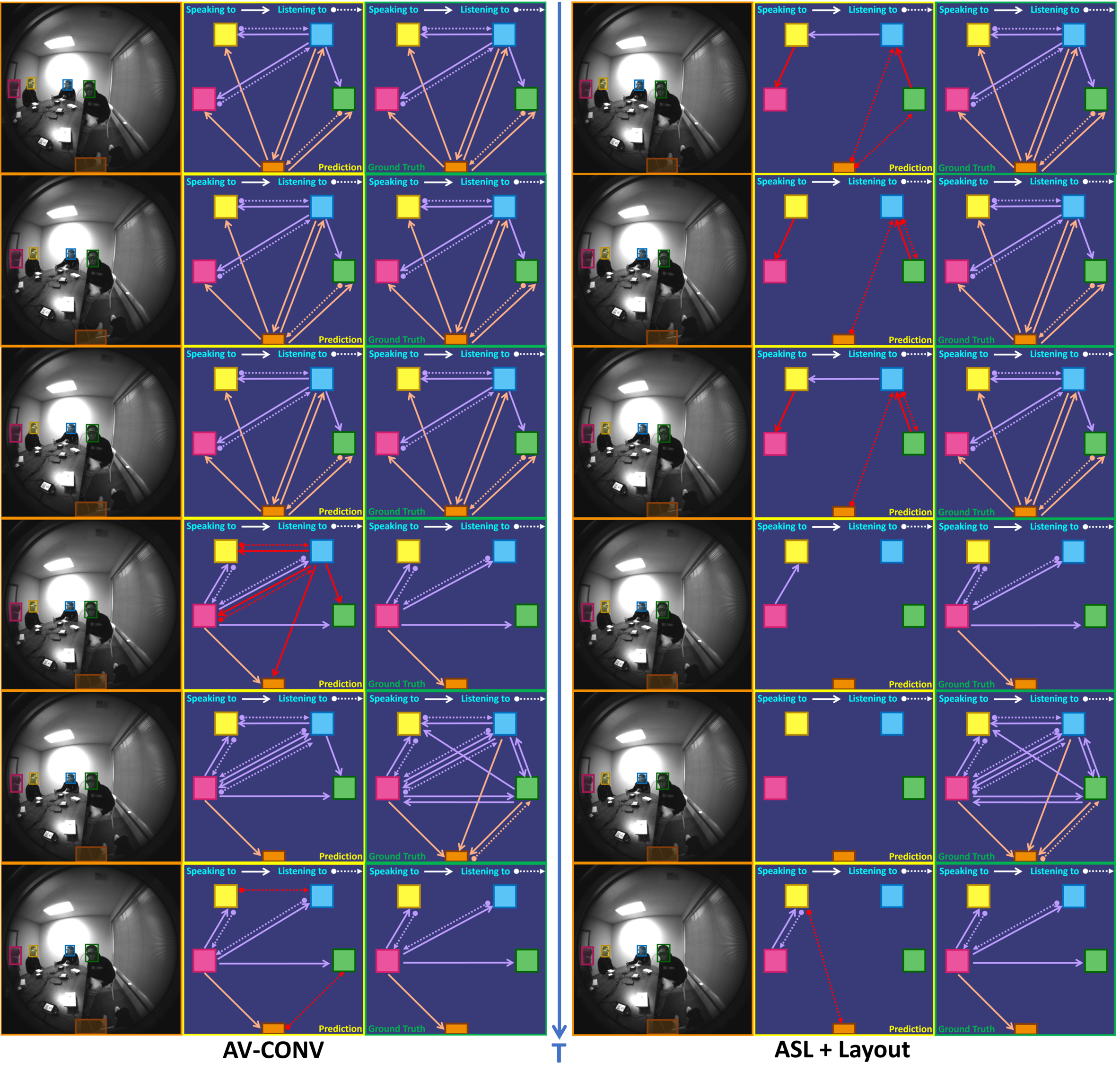}
\vspace{-1.5em}
\captionof{figure}{For (d) in Fig.~\ref{fig:suppl_viz_2}, we additionally provide visualization with the prediction results using the ASL+Layout baseline to demonstrate the superiority of our \name model.}
\label{fig:asl_viz}
\end{figure*}

In Fig.~\ref{fig:asl_viz}, we provide qualitative results from ASL+Layout baseline of Fig.~\ref{fig:suppl_viz_2}(d). 

\section{Limitation and Future Work.}
\label{sec:limit}
Our current available dataset does not include any complex group dynamics such as free-standing or walking scenario, or splitting and merging behaviors of conversational groups. With additional efforts on generating annotations, our task can further extend to large-scale dataset like Ego4D. It is also possible to include Natural language processing (NLP) module for context and intention detection in a concurrent conversational setting. 


\section{Demo Video}
\label{sec:video}
We include videos featuring demonstrations of ~\ref{sec:quali} together with the \textit{Egocentric Concurrent Conversations Dataset} and the code on our \href{https://vjwq.github.io/AV-CONV/}{AV-CONV project page}.

\begin{figure*}[t]
\centering
\includegraphics[width=1.0\linewidth]{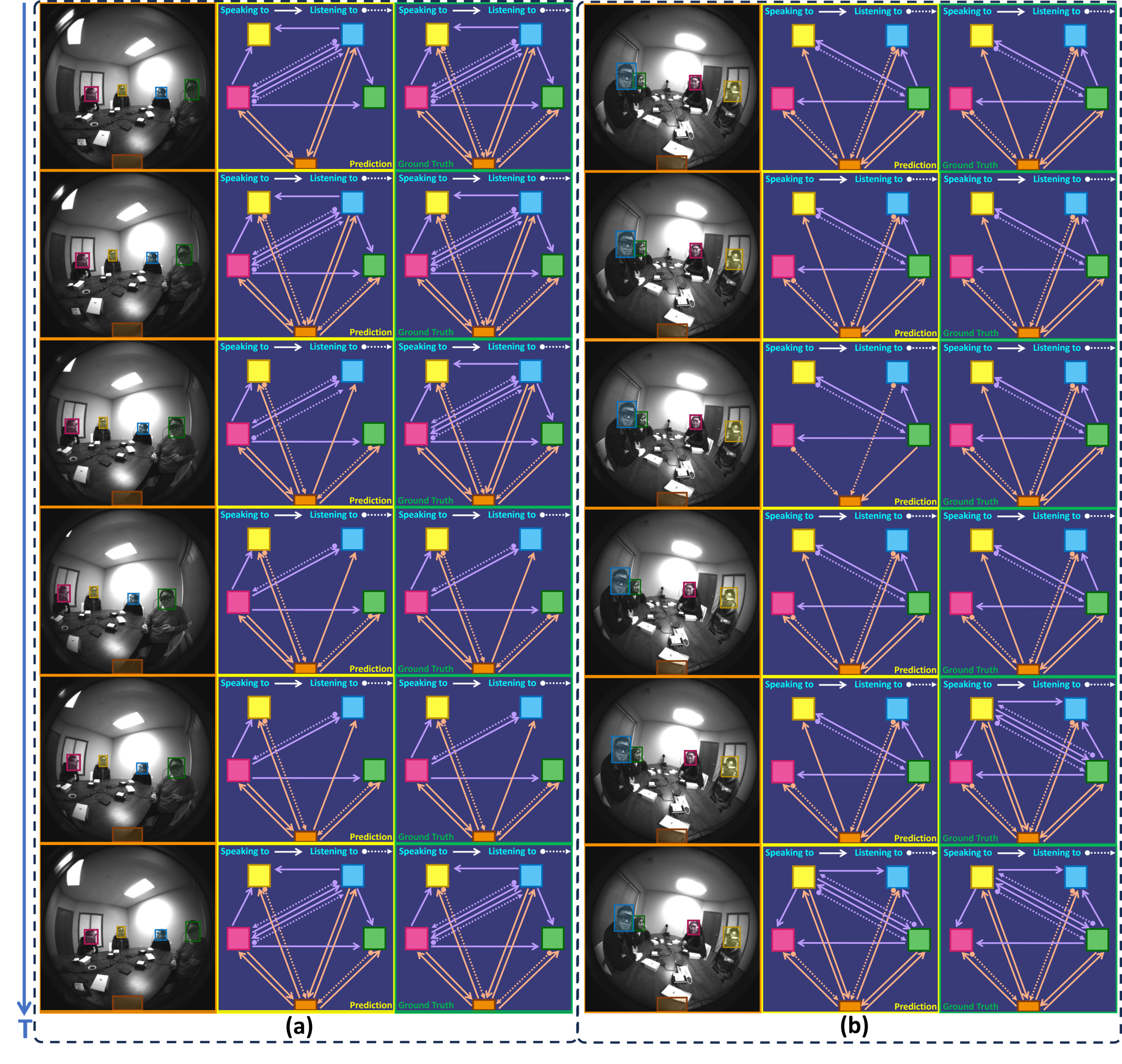}
\end{figure*}

\end{document}